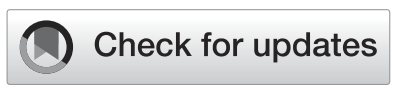

*CORRESPONDENCE
Matthias Mayer,
✉ matthias.mayer@tum.de

# Smart placement, faster robots—a comparison of algorithms for robot base-pose optimization

Matthias Mayer* and Matthias Althoff

Chair of Robotics, Artificial Intelligence and Real-Time Systems, TUM School of Computation, Information and Technology, Technical University of Munich, Munich, Germany

Robotic automation is a key technology that increases the efficiency and flexibility of manufacturing processes. However, one of the challenges in deploying robots in novel environments is finding the optimal base pose for the robot, which affects its reachability and deployment cost. Yet, existing research on automatically optimizing the base pose of robots has not been compared. We address this problem by optimizing the base pose of industrial robots with Bayesian optimization (BO), exhaustive search (ES), genetic algorithms (GAs), and stochastic gradient descent (SGD), and we find that all algorithms can reduce the cycle time for various evaluated tasks in synthetic and real-world environments. Stochastic gradient descent shows superior performance with regard to the success rate, solving more than 90% of our real-world tasks, while genetic algorithms show the lowest final costs. All benchmarks and implemented methods are available as baselines against which novel approaches can be compared.

## 1 Introduction

An often overlooked factor in the deployment of robots is the positioning of their base, which can substantially improve their performance, usually without additional monetary cost. The trend toward lighter robots also supports a more flexible base-pose selection as these robots need less sturdy foundations. By optimizing the placement of the robot, we can unlock hardware savings by making a robot more productive or by making it possible to use a simpler (modular) robot.

In this paper, we compare algorithms that optimize the base pose of robots to improve their productivity. We focus on methods that do not require significant adaptations to a specific robot, such that they can, in principle, accommodate changing modular robots. Within the literature, we found exhaustive search (ES) (Lechler et al., 2021), gradient-based methods (Son and Kwon, 2019), genetic algorithms (GAs) (Mitsi et al., 2008), and Bayesian optimization (BO) (Kim et al., 2021) as the most used methods to optimize robotic base poses. To apply gradient-based methods to arbitrary robots, we adapted Adam (Kingma and Ba, 2015), a stochastic gradient descent (SGD) method, to base placement optimization. Compared to the study by Son and Kwon (2019), which assumes a spherical wrist, our SGD only requires a well-defined forward kinematics.

Our evaluation reveals that SGD succeeds significantly more often than other methods in the tasks with the most goals. In contrast, GA significantly lowers cycle times in tasks that force the robot to be placed in one part of the optimization domain, as shown in Figure 1b.

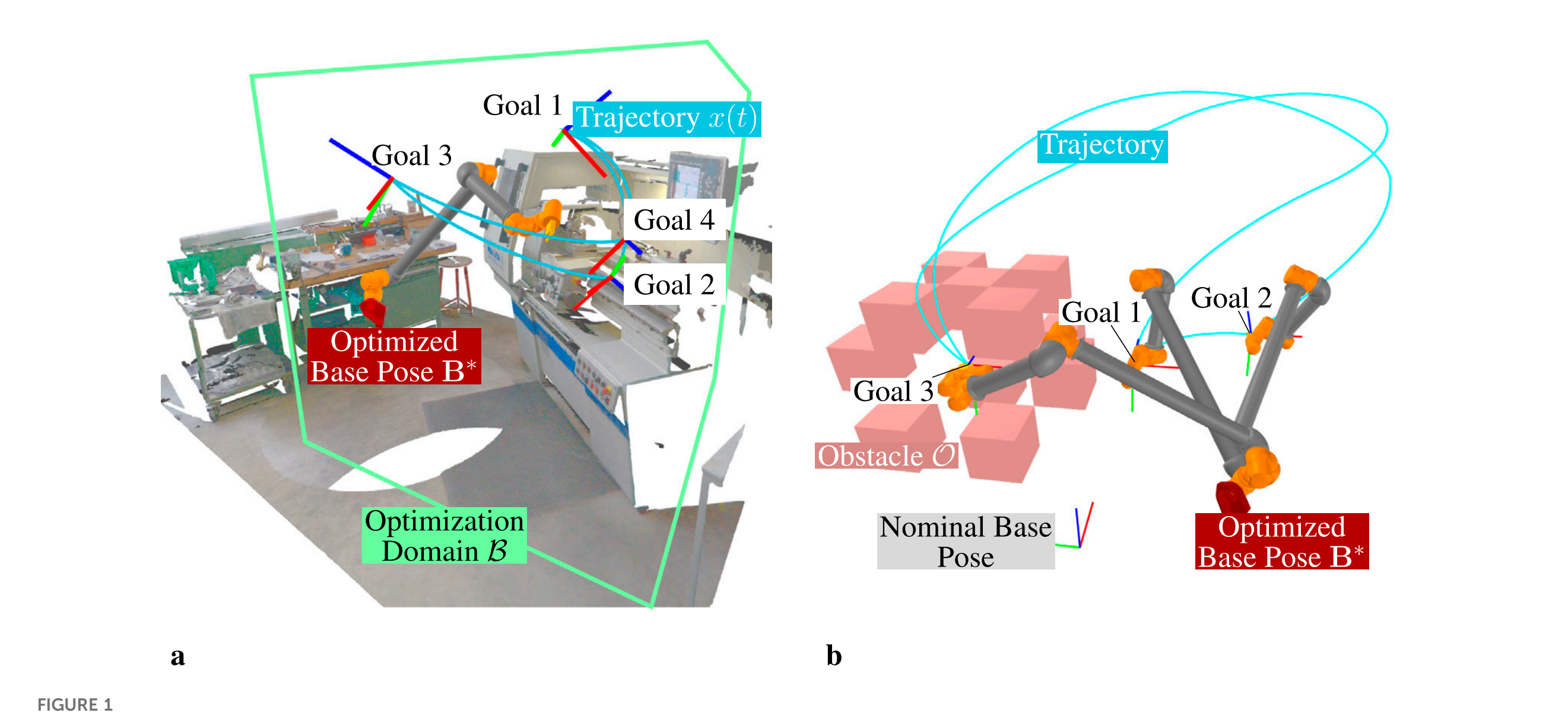

FIGURE 1
Visualization of the optimized base pose **B*** of robot[6] $R$ solving point-to-point movements in two different environments. The trajectory $x(t)$ connects the goal poses drawn as coordinate frames. On the left, we indicate the allowed base positions of the robot $\mathcal{B}$ in 3D space by the green outline of a box. Additional information, such as animations and all other best solutions, is available at https://cobra.cps.cit.tum.de/tools/rbo. **(a)** Robot solving the task inside the 3D scan. **(b)** Robot solving the task in the edge set.

In most cases, BO is significantly outperformed, having the lowest success rate and the highest average cycle times. Additionally, we find strong generalization abilities for the fine-tuned algorithms, particularly for the task set containing 3D-scanned environments, as shown in Figure 1a.

## 1.1 Related work

We summarize the state of the art for base placement optimization and discuss related and established benchmarks. Optimizing the base pose can be considered part of laying out a manufacturing system. Moslemipour et al. (2012) discussed different methods, mostly from derivative-free black-box optimization.

Noticeable exceptions are Son and Kwon (2019) and Baumgärtner et al. (2024), which formulate base placement optimization as (convex) optimization problems. Son and Kwon (2019) exploited the fact that the inverse kinematics of industrial robots with spherical wrists can be decomposed into a positioning and orientation sub-problem. Baumgärtner et al. (2024) optimized the entire robot structure, but particularly the robot base, jointly in a trajectory tracking problem. They modeled the base as a set of zero-velocity joints and solved the whole optimization via collocation.

We group the remaining methods into exhaustive search, capability maps, genetic algorithms, and Bayesian optimization.

a. ES: This method optimizes the base position of an industrial robot by evaluating all base positions on an equidistant grid (Lechler et al., 2021).
b. Capability maps: Another approach is capability maps, as used by Boschetti et al. (2013), Lin et al. (2017), and Makhal and Goins (2018). All of these exploit rather expensive pre-calculations of capability maps for the considered robots, e.g., discretizing the robot workspace into voxels and storing the manipulability at the center of each voxel. These approaches position the robot such that all desired workspace poses lie within the areas of high capability.
c. GAs: Mitsi et al. (2008) used genetic algorithms to jointly optimize the base position and initial guesses for numerical inverse kinematic solutions. The initial inverse kinematic guesses are refined by local search using a quasi-Newton algorithm. A fitness function evaluates the closeness to all desired poses and maximizes manipulability at the desired poses.
Recently, base poses and modular reconfigurable robots have been jointly optimized by Romiti et al. (2023), Lei et al. (2024), and Mayer and Althoff (2025). They encoded the base pose as up to six additional discretized genes for position and orientation and prepended them to the robot module encoding.
d. BO: Kim et al. (2021) used BO to optimize the shoulder placement of the arms of a bimanual humanoid robot. The authors particularly highlighted the ability to explore and exploit the Pareto front of optimal base poses with regard to a set of different manipulation tasks the robot should fulfill.
e. Benchmarks: The previous base placement optimization algorithms have not yet been compared on a common set of benchmark tasks. In the robotics community, various benchmark suites have helped select promising algorithms for many related problems. One example is motion planning, where Table 1 in Chamzas et al. (2022) provided a great overview of available benchmarks. Moreover, in the area of grasp planning, various benchmark sets and challenges have been proposed, as summarized by Mahler et al. (2019) (p. 1).

Within the area of robotic assembly, challenges such as those run by NIST[1] have helped make progress in tasks such as the insertion of various parts and the handling of limp objects, e.g., wires and belts. The found benchmarks either excluded the robot base position from the optimization problem (Chamzas et al., 2022; Mahler et al., 2019), focusing on the motion/grasp planning problem, or left the whole system design open (NIST), such that the base optimization cannot be analyzed on its own. To isolate the base optimization, we use the benchmark suite CoBRA (Mayer et al., 2024), which describes various robotic tasks, including motion goals, obstacles to move around, and constraints to obey. Specifcally, we rely on CoBRA to define the optimization domain $\mathcal{B}$ for the robot base and to provide (modular) robots to solve our tasks with. These found solutions can be executed on real robots, as shown by Boschetti et al. (2013), Lin et al. (2017), Makhal and Goins (2018), Romiti et al. (2023), Lei et al. (2024), and Mayer and Althoff (2025).

We focus on robotic tasks that require point-to-point movements, as can be found, e.g., in machine tending, (spot) welding, or dispensing, which were used by Lechler et al. (2021), Boschetti et al. (2013), Makhal and Goins (2018) Mitsi et al. (2008), and Kim et al. (2021). According to the IFR International Federation of Robotics (2024), these still constitute the majority of tasks automated by robots, ensuring the broad applicability of our approach.

## 1.2 Contributions

We present the first work comparing base placement optimization algorithms on a set of benchmarks and adapt Adam (Kingma and Ba, 2015), an SGD method, to the base placement optimization problem. In particular, we

- compare base placement optimization methods [GA, BO, random sampling (RS), and SGD] with respect to cost convergence and success rate;
- propose benchmarks for base placement optimization;
- for the first time, apply SGD base placement optimization;
- test the generalizability of base placement optimization methods to various tasks and with different allowed base poses;
- systematically tune the hyperparameters of GA, BO, and SGD for base placement optimization.

In Section 2, we define the optimization problem and introduce the evaluated methods. Section 3 states the setup for our evaluation and summarizes the results. These are discussed in Section 4, which is followed by a conclusion in Section 5.

1 www.nist.gov/el/intelligent-systems-division-73500/robotic-grasping-and-manipulation-assembly/assembly, accessed on 5th November 2025.

# 2 Methods

This section defines the base placement optimization problem and presents how the considered optimization methods can solve it. We closely follow the notation introduced in CoBRA (Mayer et al., 2024). In summary, we use lowercase letters for scalars, e.g., $t_f \in \mathbb{R}$; bars to indicate vectors, e.g., $\bar{b} \in \mathbb{R}^N$; uppercase and bold letters for poses, e.g., $\mathbf{B}$; and calligraphic letters for sets, e.g., $\mathcal{G}$.

Our robotic tasks are given as a tuple containing a set of goal poses for the end effector $\mathcal{G}$, a set of constraint functions $\mathcal{C}$, a set of environment obstacles $\mathcal{O}$, and a cost function $J_\mathrm{C}$. The set of constraints $\mathcal{C}$ enforces common desired properties, such as respecting the joint limits and avoiding collisions with $\mathcal{O}$. For each task, a trajectory $x(t)$ and base pose $\mathbf{B} \in SE(3)$, i.e., a position and orientation[2], need to be found that satisfy all constraints in $\mathcal{C}$. The trajectory needs to pass through all goals $\mathcal{G}$ during its execution time $t_f$, as described in Section 2.5.

Our paper aims to find the optimal base pose

$$\mathbf{B}^* = \arg\min_{\mathbf{B}} \left( J_\mathrm{C}\left(R, \mathbf{B}, x(t)\right) \right) \tag{1}$$

subject to

$$\forall t \in [0, t_f]\ \forall c \in \mathcal{C}\colon\ c\left(x(t), t, \mathbf{B}, R\right) \le 0. \tag{2}$$

For notational convenience, we additionally introduce the set of base poses that fulfill all constraints:

$$\mathcal{B} = \{\mathbf{B} \in SE(3) \mid \forall c \in \mathcal{C}\colon\ c\left(x(t), t, \mathbf{B}, R\right) \le 0\}. \tag{3}$$

For all considered optimizers, we need an $a$-dimensional vector $\bar{b} \in \mathbb{R}^a$ to parameterize $\mathbf{B}$, where $\bar{b}_{i:j} = [b_i, \ldots, b_j]$ contains the elements $i$ through $j$ from $\bar{b} = [b_1, \ldots, b_N]$ and $\bar{0}_N$ is the n-dimensional vector of zeros. We use

- $a = 3$: Only the Cartesian position is used, e.g., in Lechler et al. (2021):

$$\mathbf{B}_1\left(\bar{b}\right) = \begin{bmatrix} \mathbf{1}_{3\times3} & \bar{b}_{1:3} \\ \bar{0}_3^T & 1 \end{bmatrix}. \tag{4}$$

- $a = 6$: Adds an axis angle encoding (Siciliano and Khatib, (2016), Table 2.1), similar to Romiti et al. (2023), and extending Mitsi et al. (2008) and Kim et al. (2021):

$$\mathbf{B}_2\left(\bar{b}\right) = \mathbf{B}_\mathrm{pos}\left(\bar{b}_{1:3}\right) \begin{bmatrix} \mathbf{R}_\mathrm{axis-angle}\left(\bar{b}_{4:6}\right) & \bar{0}_3 \\ \bar{0}_3^T & 1 \end{bmatrix}. \tag{5}$$

Next, we present the considered optimization approaches.

2 $SE(3) = \mathbb{R}^3 \times SO(3)$ is the space representing an arbitrary pose, i.e., position $\mathbb{R}^3$ and orientation $SO(3)$ in three-dimensional space. Commonly represented as a $4 \times 4$ homogeneous transformation matrix (Siciliano and Khatib, 2016; Section 2.2.3).

## 2.1 Baseline and exhaustive search

As a baseline, we implement a dummy base placement optimization that keeps the nominal pose of the robotic task. Additionally, we consider an RS optimizer that uniformly samples poses from the allowed set $\mathcal{B}$. With an increasing number of sampled points, it approximates the ES method surveyed in Section 1.1.

## 2.2 Genetic algorithms

GAs belong to the set of black-box optimization algorithms. Within a GA, a population of individual solution candidates encoded as a list of genes is optimized. Those genes are our parameter vectors, defined in Equations 4, 5, and are altered by

- mutation, which locally alters a single gene or several genes of an individual;
- cross-over, which combines genes from two individuals to create a new "offspring";
- selection, which determines how individuals are added to the next generation.

The mutation operator uniformly selects single values $b_i, i \in [1, \ldots, a]$ to alter and adds a random number from a scalar normal distribution with zero mean and unit standard deviation $\mathcal{N}(0,1)$ to them. Cross-over selects the first $n \in [1, \ldots, a]$ entries in one gene vector and combines it with the last $a - n$ elements from another individual. The selection is based on the negative cost function $-J_C(\mathbf{B})$, which is considered the fitness of each individual.

## 2.3 Bayesian optimization

BO is another common method within derivative-free or black-box optimization algorithms. Similar to Kim et al. (2021), our BO uses a Gaussian process $\hat{J}_C(\mathbf{B})$ to approximate the cost of $i$ previously tested base poses $\{J_C(\mathbf{B}_1), \ldots, J_C(\mathbf{B}_i)\}$. The next base pose $\mathbf{B}_{i+1}$ to explore is sampled from $\hat{J}_C(\mathbf{B})$ based on the exploration/exploitation hyperparameter $\xi$. The resulting cost $J_C(\mathbf{B}_{i+1})$ can, in turn, be used to refine $\hat{J}_C(\mathbf{B})$. For base placement optimization, we define $\hat{J}_C(\mathbf{B})$ using $\mathbf{B}(\bar{b}) \in \mathbb{R}^a$ following Equations 4, 5 and use it to fit the scalar cost value $J_C(\mathbf{B})$. At the start, BO commonly samples $n_{\text{init}} \in \mathbb{N}_{>0}$ points, e.g., randomly or from low-discrepancy sequences, such as the Hammersley set (Wong et al., 1997). The initial sampling method is also determined via hyperparameter tuning.

## 2.4 Stochastic gradient decent

SGD is a gradient-based method that gained prominence for training deep neural networks. Here, it serves as the outer method of a two-level optimization. A current state-of-the-art SGD algorithm is Adam (Kingma and Ba, 2015). Given a manipulator with known forward kinematics

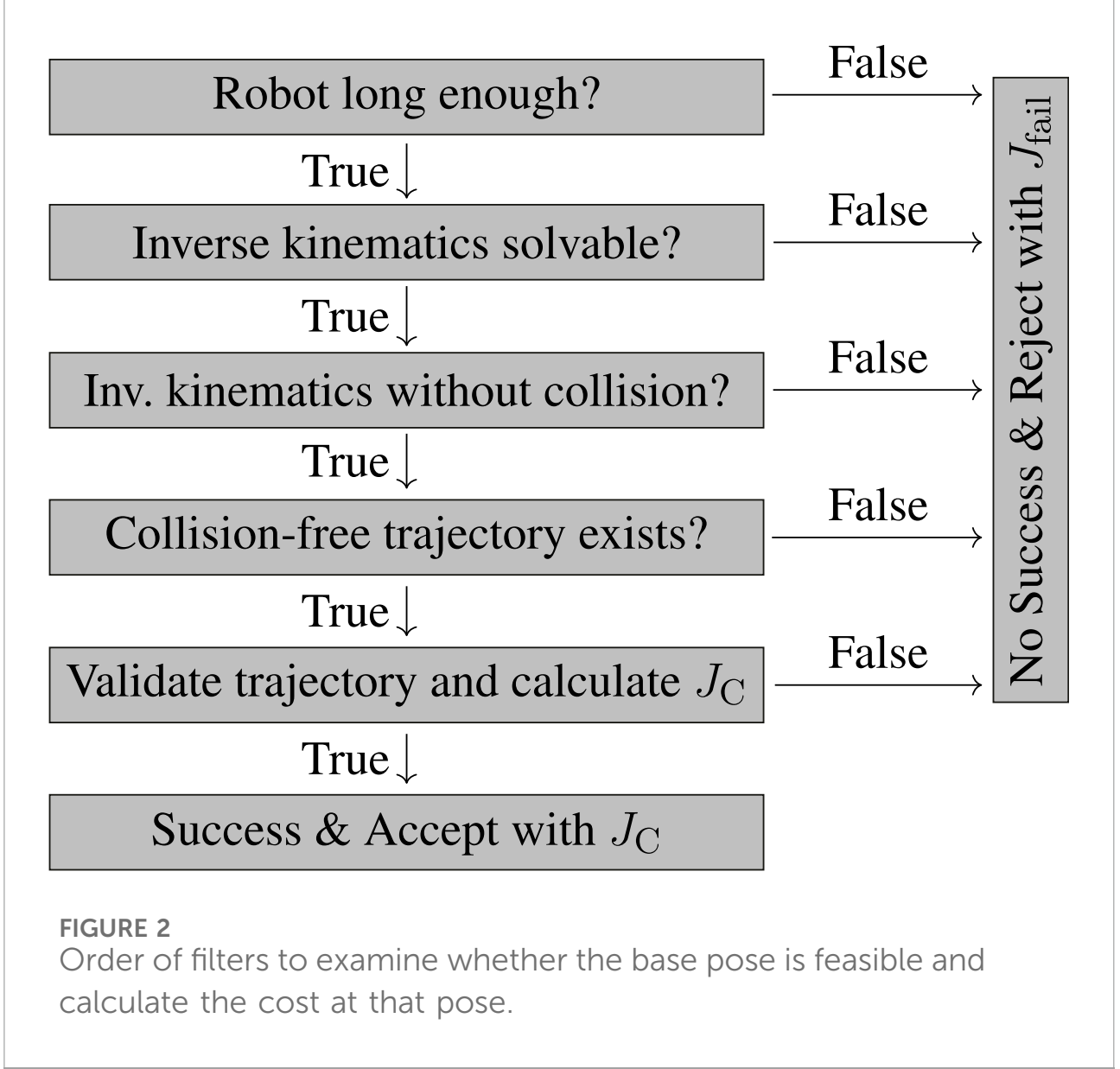

FIGURE 2
Order of filters to examine whether the base pose is feasible and calculate the cost at that pose.

FK: $\mathbb{R}^{n_{\text{DoF}}} \to SE(3)$, i.e., a function returning the end-effector pose of a robot given its joint angles $\bar{q} \in \mathbb{R}^{n_{\text{DoF}}}$ and a distance function $\delta$: $SE(3) \times SE(3) \to \mathbb{R}_{\geq 0}$, we can use numerical inverse kinematics (IK) to find configurations $\bar{q}_i$ close to desired goal poses $g_i \in SE(3)$ (Siciliano and Khatib, 2016; Section 2.7) (inner method). We use Adam to optimize the base pose, such that $\delta$ is reduced for the found configurations $\bar{q}_i$. Starting at a random base pose determined by $\bar{b} \in \mathbb{R}^a$, we run the following two-level optimization loop $n_{\text{Adam}} \in \mathbb{N}_{>0}$ times.

1. We move the robot to $\mathbf{B}(\bar{b})$ and find the closest IK solutions $\bar{q}_i \in \mathbb{R}^{n_{\text{DoF}}}$ to all goals $g_i \in \mathcal{G}$ according to a distance function (Equation 6)

$$\delta\left(\mathbf{B}(\bar{b})\text{FK}(\bar{q}_i), g_i\right) \geq 0, \tag{6}$$

within a maximum of $n_{\text{IK}} \in \mathbb{N}_{>0}$ steps. We terminate if all IK solutions are within the tolerance of each goal.

2. We calculate the gradient of $\delta$ with respect to the base parameters $\bar{b}$ at each found IK solution $\bar{q}_i$ (Equation 7):

$$\nabla\delta_i = \frac{\partial\delta\left(\mathbf{B}(\bar{b})\text{FK}(\bar{q}_i), g_i\right)}{\partial\bar{b}}. \tag{7}$$

We then average them as $\nabla\delta = \sum_{i=0}^{|\mathcal{G}|}\nabla\delta_i$, and we apply the Adam step to $\bar{b}$ based on an exponential decaying average of $\nabla\delta$ determined by $\beta_1, \beta_2 \in [0,1)$ and a step size $\alpha > 0$ (Kingma and Ba, 2015, Alg. 1).

The algorithm is restarted from other random initial guesses while there is time left to escape local minima.

## 2.5 Task solver

All base placement optimization methods have access to the same task solver that judges whether a suggested base pose $\mathbf{B}$

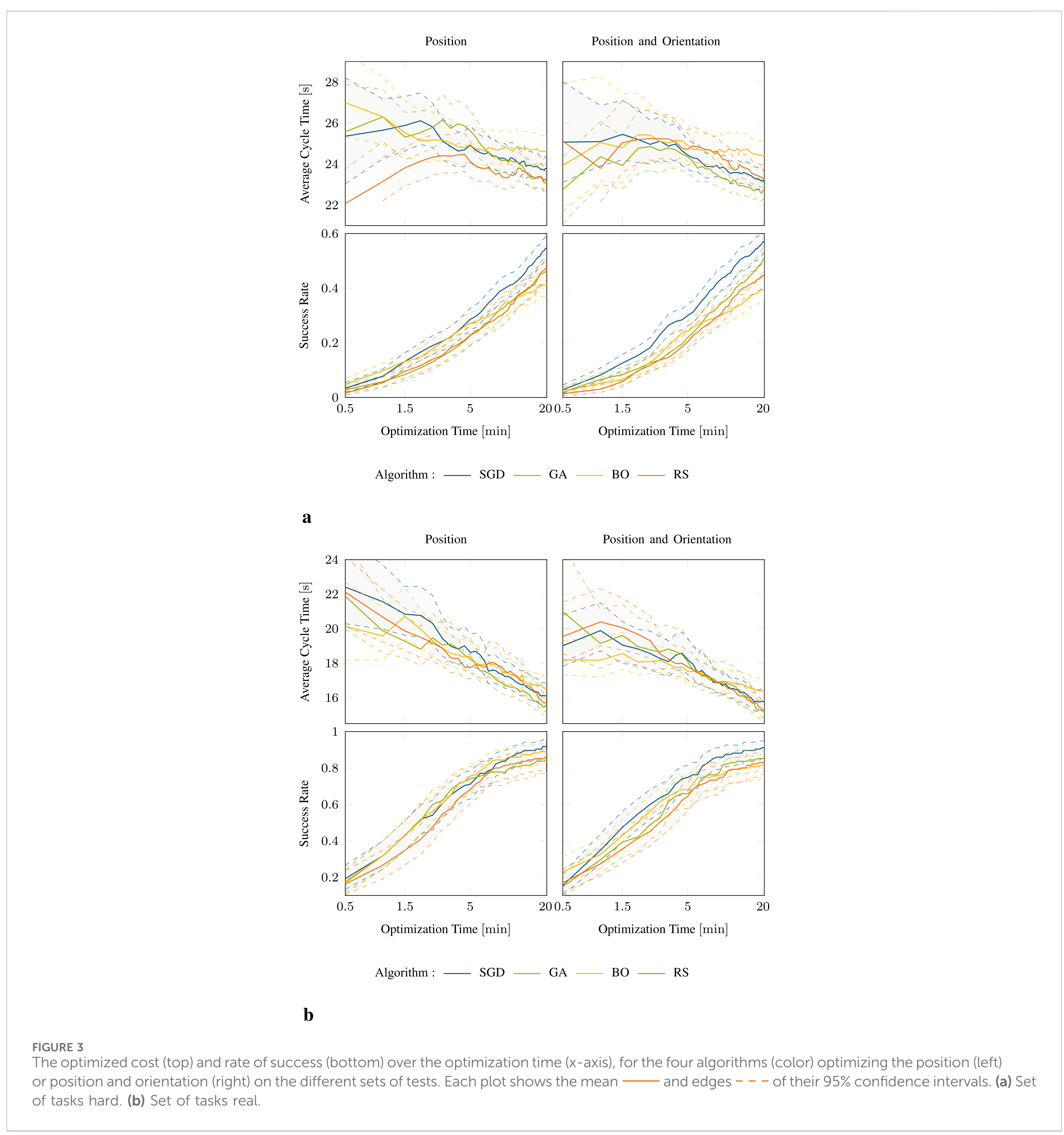

FIGURE 3
The optimized cost (top) and rate of success (bottom) over the optimization time (x-axis), for the four algorithms (color) optimizing the position (left) or position and orientation (right) on the different sets of tests. Each plot shows the mean ——— and edges – – – of their 95% confidence intervals. **(a)** Set of tasks hard. **(b)** Set of tasks real.

allows the robot $R$ to fulfill all goals $\mathcal{G}$ of a given task. This solver first applies step-by-step elimination (Althoff et al., 2019, p. 7–8) to reject base poses further away from any goal than the maximum length of the robot or those without (collision-free) inverse kinematic solutions. If these filters pass, we run RRT-Connect (Kuffner and La Valle, 2000) and the trajectory generator proposed by Kunz and Stilman (2012) to find a collision-free and feasible trajectory $x(t)$ for the robot positioned at $\mathbf{B}$. The overall process is summarized in Figure 2. We calculate the cost function $J_C$ based on the found trajectory. If any step fails, a fixed failure cost $J_{fail} \in \mathbb{R}$ is returned to the optimizer.

The overall runtime of the task solver varies depending on which filter rejects the suggested base pose. For reference, a failure in the first two steps can be determined within milliseconds as these only need a few forward kinematic calculations. Collision-free inverse kinematic solutions are identified in approximately 0.1 s, while the path planner succeeds in the order of seconds and only fails after a fixed time-out is reached.

# 3 Numerical experiments

We ran the following numerical experiments to evaluate and compare the suggested base placement optimization algorithms[3]. First, we present the setup, including the set of considered robotic tasks and specific libraries for implementing the optimizers. Second, we analyze the adaptability, convergence, and generalizability of each approach for various robotic tasks.

## 3.1 Setup

In all our experiments, we use Timor Python (Külz et al., 2023) for kinematic calculations and collision detection. Our benchmark tasks are implemented as CoBRA tasks (Mayer et al., 2024). We created our simple set of 100 tasks by sampling three cubic obstacles on a grid around the origin and three desired poses outside those cubes that the end effector should stop at, as proposed by Whitman et al. (2020). Moreover, we tested the tuned methods on the following sets:

- hard: 100 synthetic tasks using the same sampling as the simple set of tasks, with five goals and obstacles instead of three.
- real: 27 tasks based on 3D-scanned real-world factory settings in which CNC machines are fed, e.g., Figure 1a. As can be observed, these examine how methods react to non-cubic and irregular obstacles.
- edge: 100 particularly created cases where goals are all located outside $\mathcal{B}$ and additionally placed close to obstacles to limit the feasible base poses, such as goal 3 in Figure 1b.

The tasks[4] are tagged with BPO24 and simple/hard/real/edge_case_hard in CoBRA.

All tasks have the following constraints:

- Joints limited in position $q$ and torque $\tau$,
- no (self-)collisions,
- all goal poses need to be passed, and
- the robot base must be positioned around a given pose with ±1 m in each Cartesian direction and unrestricted orientation[5].

In all cases, the goal was to minimize the cycle time $J_{\mathrm{T}} = t_{\mathrm{f}}$ required to fulfill the given task with a given robot[6]. This objective has been shown to work for deployments in real-world tasks and is positively correlated with other costs, such as energy consumption (Mayer and Althoff, 2025). If the solver cannot find a solution trajectory for a suggested base pose, a default failure cost of $J_{\mathrm{fail}}$ =20 s is returned, which is well above the possible cycle time. All experiments were run on a 64-core AMD EPYC 7742 clocked at 2.25 GHz. The inverse kinematic solver and Equation 6 in SGD used a weighted sum of Euclidean distance and rotation angle between poses $\delta$, with the default weight 5 mm/1.8° of Timor Python (Külz et al., 2023). We, therefore, keep this weighting parameter fixed.

The SGD optimizer used Adam and automatic differentiation implemented using PyTorch 2.2.1 (Paszke et al., 2019). The GA was implemented with PyGAD (Gad, 2023). BO used the implementation provided by scikit-optimize[7]. SGD, GA, and BO require setting hyperparameters to tune them for specific applications, in contrast to dummy and RS. We used Optuna (Akiba et al., 2019) in its default setting with the tree-structured Parzen estimator (TPE) optimizer (Bergstra et al., 2011) and median pruner to minimize the mean cost on the first 70% of simple tasks. The base placement optimization was limited to finding the best position for the robot base by using Equation 4. In total, we ran 400 trials for each of the three tunable algorithms with Optuna. The optimized hyperparameters are listed in the Supplementary Material. For RS, we did not identify any tunable parameter; it always uses uniform distribution sampling from the whole domain of $\bar{b}$.

The final comparison of tuned algorithms was performed on a distinct set of tasks not used during hyperparameter optimization, i.e., the last 30% of the simple set and the complete sets hard, real, and edge. For all sets in addition to simple, we had to increase the failure penalty to $J_{\mathrm{fail}}$ =50 s as the additional goals and more complex obstacles need, on average, longer cycle times (see average cycle times in Figure 3). The comparison with tasks outside the simple set is particularly important as it shows the generalization potential of each algorithm. Algorithms are compared in a fixed-budget setting: they obtain $t_{\mathrm{CPU}}$ = 1200 s = 20 min[8] per task to explore possible base poses. This time limit encompasses all optimization steps, including the internal calculations for each optimization method, running the provided filters, and, upon success, completing the path planning. The final score is the cost of the best solution, and we present statistics over five distinct seeds.

## 3.2 Results

First, we present the convergence of the minimum cost found (top) and the success rate (bottom) on the hard (Figure 3a) and a real set of tasks (Figure 3b). The success rate is the fraction of tasks and seeds for which each base placement optimization algorithm can find a valid base pose within the optimization time as the x-axis. The first column of each subfigure lists the results obtained by optimizing only the base position

3 All code is available at https://gitlab.lrz.de/tum-cps/robot-base-pose-optimization

4 Available at: https://cobra.cps.cit.tum.de/tools/rbo

5 Limiting each vector element of the axis-angle representation to $[-\pi, \pi]$ results in a bounded search space covering all orientations. These limits are enforced in the SGD and GA by clipping after applying the gradient step or the genetic operators.

6 Made from the CoBRA module set modrob-gen2 with module order $M$ = [105, 2, 2, 24, 2, 25, 1, 1, 1, GEP2010IL] and shown in Figures 1a,b.

7 https://scikit-optimize.github.io/stable/modules/generated/skopt.Optimizer.html, accessed on 5th November 2025.

8 Set by preliminary tests on the simple set, which showed convergence after this optimization time.

TABLE 1 The mean and 95 % confidence interval (CI) for the best found cycle time (upper half) and success rate (lower half) of each algorithm (columns). Arrows ↑, ↓ indicate the direction of better values. Highlighted means are outside all other CIs on the *worse* or better side.

| Measurement | Action space | Task set | RS | | | GA | | | BO | | | SGD | | |
|---|---|---|---|---|---|---|---|---|---|---|---|---|---|---|
| | | | Mean | CI | | Mean | CI | | Mean | CI | | Mean | CI | |
| ↓ Average best cycle time [s] | Position | Simple | 10.77 | 10.50 | 11.15 | 10.69 | 10.41 | 11.08 | *11.37* | 11.10 | 11.73 | 10.81 | 10.54 | 11.20 |
| | | Hard | 23.29 | 22.78 | 23.93 | 23.09 | 22.64 | 23.64 | *24.65* | 23.98 | 25.42 | 23.77 | 23.26 | 24.36 |
| | | Real | 15.70 | 15.20 | 16.24 | 15.53 | 15.05 | 16.08 | 16.55 | 15.95 | 17.24 | 16.11 | 15.54 | 16.86 |
| | | Edge | 15.97 | 15.11 | 17.14 | **13.65** | 12.98 | 15.44 | *18.13* | 16.23 | 22.02 | 16.26 | 15.28 | 17.48 |
| | Position + rotation | Simple | 10.84 | 10.57 | 11.22 | 10.67 | 10.42 | 11.08 | *11.26* | 10.97 | 11.64 | 10.74 | 10.47 | 11.13 |
| | | Hard | 23.28 | 22.80 | 23.87 | 22.71 | 22.30 | 23.26 | *24.40* | 23.78 | 25.09 | 23.11 | 22.66 | 23.63 |
| | | Real | 15.28 | 14.86 | 15.82 | 15.22 | 14.75 | 15.77 | 16.38 | 15.85 | 17.04 | 15.74 | 15.19 | 16.47 |
| | | Edge | 15.64 | 15.16 | 16.21 | **14.28** | 13.90 | 14.70 | *17.61* | 16.78 | 18.65 | 16.34 | 15.78 | 17.00 |
| ↑ Success rate [%] | Position | Simple | 96.67 | 92.67 | 98.67 | 96.67 | 92.67 | 98.67 | 96.67 | 92.67 | 98.67 | 96.67 | 92.67 | 98.67 |
| | | Hard | 47.40 | 43.00 | 51.80 | 46.40 | 42.00 | 50.80 | *41.40* | 37.20 | 45.80 | **54.60** | 50.20 | 59.00 |
| | | Real | 85.93 | 79.26 | 91.11 | 84.44 | 77.78 | 89.63 | 88.89 | 82.96 | 93.33 | 91.85 | 85.93 | 95.56 |
| | | Edge | 14.60 | 11.80 | 18.00 | 13.80 | 11.00 | 17.00 | *5.40* | 3.60 | 7.60 | 15.00 | 12.00 | 18.40 |
| | Position + rotation | Simple | 96.67 | 92.67 | 98.67 | 96.67 | 92.67 | 98.67 | 96.67 | 92.67 | 98.67 | 96.67 | 92.67 | 98.67 |
| | | Hard | 44.80 | 40.40 | 49.00 | 50.60 | 46.20 | 55.00 | *39.40* | 35.20 | 43.60 | **57.00** | 52.80 | 61.20 |
| | | Real | 82.96 | 76.30 | 88.89 | 85.19 | 78.52 | 90.37 | 81.48 | 74.07 | 87.41 | **91.11** | 85.19 | 94.81 |
| | | Edge | 54.40 | 50.20 | 58.67 | 55.20 | 50.80 | 59.60 | *25.00* | 21.40 | 29.00 | 55.40 | 51.00 | 59.80 |

(see Equation 4) used during hyperparameter optimization; the second column lists the results when optimizing the base position and orientation (see Equation 5). The mean and 95% confidence interval calculated via bootstrapping over all tasks and seeds are shown for each method.

Second, we summarize the final average best cycle time and the success rate of each method across all sets of tasks in Table 1. To judge significance, we provide the mean expected cost/success rate over all found solutions, along with 95% confidence intervals calculated using bootstrapping. Therefore, the numbers provided in Table 1 for the hard task set show the mean and confidence interval at $t_{CPU}$ from Figure 3. We highlight mean values outside the confidence interval of all other methods in the same row as worse or **better**. Furthermore, we conduct a Wilcoxon signed-rank test with Bonferroni $p$-value corrections over all tasks used in the evaluation to identify significant differences between algorithms; the full test statistics are provided in Supplementary Tables S4, 5. The hard, real, and edge sets of tasks test generalization as they have different distributions of goal poses and obstacles compared to the simple set of tasks used for hyperparameter optimization. We omit the dummy optimizer as it can only solve the simple set of tasks in 47% of cases and the hard set of tasks in 18% while failing to solve any task from the other sets.

# 4 Discussion

With regard to overall performance, we find that all algorithms outperform the dummy optimizer, highlighting that base-pose optimization reduces cycle time. Moreover, all methods constantly show noticeable confidence intervals in Figure 3 and Table 1 due to the stochasticity of the evaluated approaches and the underlying trajectory generation process (see Section 2.5). The time limit of 1200 s seems sufficient for the simple and real sets of tasks, as evidenced by the success rate plateauing in Figure 3b. The edge and hard sets of tasks could improve with more computation time as their success rates have not leveled off. Referencing the Wilcoxon tests in Supplementary Tables S4, 5, we find that, over all tasks,

- SGD significantly increases the success rate over all algorithms ($3.1 \times 10^{-71} < p < 7.5 \times 10^{-3}$),
- GA significantly decreases costs over RS and BO ($5.7 \times 10^{-108} < p < 4.8 \times 10^{-6}$), and
- BO significantly increases costs over all other algorithms ($5.7 \times 10^{-108} < p < 3.2 \times 10^{-92}$).

The state-of-the-art methods (GA, BO, and RS) perform quite similarly on the set of hard and real tasks and in both optimization domains, as shown in Figure 3. However, in the end, BO often performs worse than the other methods, as indicated by the italic means outside the other confidence intervals in Table 1. Based on the average best cost at the end of the runtime, GA seems to have a slight edge over existing methods, with lower costs on the set edge (**bold** in Table 1, upper half) and the lowest average in all other cases. With regard to the success rate, our adaptation of SGD is better in three settings (**bold** in Table 1 lower half) and shows the highest average in the other settings. It also increases the success rate faster than the other methods in three out of four cases, as shown in Figure 3.

Concerning generalization to sets of tasks not used for hyperparameter parameterization, all algorithms experienced losses in the success rate, i.e., the mean success rates in each column are outside the confidence interval for the set simple. Still, for the set real, we find success rates of more than 80% for all algorithms. On average, SGD is the most successful on the real set of tasks, with more than a 90% success rate, an improvement over RS and GA.

The benefits of the larger optimization domain for position and orientation are not realized by all methods across all sets of tasks. Considering the success rate in Table 1, we observe that RS and BO deteriorate in the sets hard and real, while GA improves its performance. Additionally, Table 1 shows that for the set edge, the larger search space increases the success rate for all methods. Finally, the larger search space also often decreases final cycle time, e.g., in three of four cases for SGD and GA.

# 5 Conclusion

For the first time, this paper compares existing robot base-pose optimizers, namely BO, RS, and GAs, with a focus on methods that do not require pre-calculations. In addition, for the first time, we apply Adam, an SGD-based optimizer, to this problem. Each algorithm was implemented in an anytime fashion, allowing the user to set a time budget for finding the best base pose. The hyperparameters of each method were optimized, and the methods were compared on a set of diverse robotic tasks, both synthetic and based on real-world 3D scans.

The three previously applied methods (BO, RS, and GA) showed comparable success rates on our set of simple tasks, but for the considered optimization time, BO found significantly worse cycle times, and GA achieved, on average, the best cycle times. The optimizer SGD showed a significantly improved success rate. It particularly outperformed the other methods in complex robotic tasks within cluttered real-world environments or tasks with more goals and obstacles. In the tasks based on real-world 3D scans, more than 90% could be solved with SGD, which outperformed the other methods. Additionally, allowing any orientation increases the success rate in our novel set of task edges, indicating that additional mounting effort benefits more complex tasks.

# Data availability statement

The code of the solvers and experiments is available at https://gitlab.lrz.de/tum-cps/robot-base-pose-optimization. The datasets containing the tasks to optimize for and generated solutions for these tasks are hosted on the project website https://cobra.cps.cit.tum.de/tools/rbo.

# Author contributions

MM: Methodology, Software, Writing – original draft, Investigation, Visualization, Data curation, Validation, Writing – review and editing. MA: Conceptualization,

Methodology, Supervision, Project administration, Resources, Funding acquisition, Writing – review and editing.

## Funding

The authors declare that financial support was received for the research and/or publication of this article. This work was supported by the Deutsche Forschungsgemeinschaft (German Research Foundation) under grant number AL 1185/31-1.

## Acknowledgements

The authors would like to thank Jonathan Külz for Timor Python and their students Friedrich Dang, Hao Gao, Lukas Seitz, Matthias Pouleau, Seok Jung, and Tom Tschigfrei who tested several related ideas.

## Conflict of interest

The authors declare that the research was conducted in the absence of any commercial or financial relationships that could be construed as a potential conflict of interest.

## Generative AI statement

The authors declare that no Generative AI was used in the creation of this manuscript.

Any alternative text (alt text) provided alongside figures in this article has been generated by Frontiers with the support of artificial intelligence and reasonable efforts have been made to ensure accuracy, including review by the authors wherever possible. If you identify any issues, please contact us.

## Publisher's note

## Supplementary material

The Supplementary Material for this article can be found online at: https://www.frontiersin.org/articles/10.3389/fmtec.2025.1642524/full#supplementary-material